\documentclass[sigconf]{aamas}

\usepackage{amsmath}
\usepackage{algorithm}
\usepackage{algpseudocode}
\usepackage{placeins}
\usepackage{booktabs,colortbl}
\definecolor{CIPblue}{HTML}{005AA7}
\definecolor{CIPink}{HTML}{172A3A}
\providecommand{\Description}[1]{}

\usepackage{balance} 

\makeatletter
\gdef\@copyrightpermission{
  \begin{minipage}{0.2\columnwidth}
   \href{https://creativecommons.org/licenses/by/4.0/}{\includegraphics[width=0.90\textwidth]{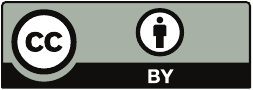}}
  \end{minipage}\hfill
  \begin{minipage}{0.8\columnwidth}
   \href{https://creativecommons.org/licenses/by/4.0/}{This work is licensed under a Creative Commons Attribution International 4.0 License.}
  \end{minipage}
  \vspace{5pt}
}
\makeatother

\setcopyright{ifaamas}
\acmConference[AAMAS 2027]{Proc.\@ of the 26th International Conference
on Autonomous Agents and Multiagent Systems (AAMAS 2027)}{3--7 May 2027}{Hanoi, Vietnam}{M.~Baldoni, F.~Fang, W.~Yeoh, N.~Yorke-Smith (eds.)}
\copyrightyear{2027}
\acmYear{2027}
\acmDOI{}
\acmISBN{}

\acmSubmissionID{1997}

\submissionType{Research Paper Track}

\title[CIPHEUR: LLM-Guided Structural Search Control]{CIPHEUR: LLM-Guided Structural Search Control for Conflict-Graph Optimization}

\author{Yuqing Cheng}
\authornote{Yuqing Cheng and Yan Jia contributed equally to this work.}
\affiliation{
	\institution{National University of Defense Technology}
	\city{Changsha}
	\country{China}}
\email{chengyuqing@nudt.edu.cn}

\author{Yan Jia}
\authornotemark[1]
\affiliation{
	\institution{National University of Defense Technology}
	\city{Changsha}
	\country{China}}
\email{jia.yan20@nudt.edu.cn}

\author{Yanjie Song}
\affiliation{
	\institution{Dalian Maritime University}
	\city{Dalian}
	\country{China}}
\email{songyj\_2017@163.com}

\author{Xiaolu Liu}
\affiliation{
	\institution{National University of Defense Technology}
	\city{Changsha}
	\country{China}}
\email{lxl\_sunny@nudt.edu.cn}

\author{Lining Xing}
\affiliation{
	\institution{Jiangsu University of Technology}
	\city{Changzhou}
	\country{China}}
\email{xln@jsut.edu.cn}

\author{Feng Yao}
\affiliation{
	\institution{National University of Defense Technology}
	\city{Changsha}
	\country{China}}
\email{yaofeng@nudt.edu.cn}

\begin{abstract}
Conflict-graph search can stagnate despite effective operators when interventions fail to address the conflicts blocking progress. A disruptive intervention may also require further search before its value emerges. We propose CIPHEUR, a large language model (LLM)-guided framework that connects conflict evidence to coordinated intervention and recovery through Structured Intervention Units (SIUs). The shared control interface supports offline synthesis of executable policies and online contextual decisions. Asynchronous online control sustains native search during inference, while all policies preserve the original objective and feasibility constraints. Across eight satellite–ground scheduling conflict views with five runs per view and a 900-second evaluation cap, CIPHEUR-online achieves the highest mean objective among 11 configurations and increases mean scheduled contact duration by 14.84\% over the greedy baseline GWMIN. It exceeds the strongest external per-view mean in seven of eight views and outperforms CHILS in all 40 paired runs. Within the same runs, its normalized objective improvement from 360 to 900 seconds is 1.50–1.86 times that of the two programmatic controllers. A separately frozen, one-shot LLM-generated controller also improves the mean objective over CHILS without runtime model calls. Together, these results demonstrate how structural search control translates LLM design knowledge and contextual reasoning into improved solution quality and sustained search progress.
\end{abstract}

\keywords{Large language models, Search control, Combinatorial optimization, Hyper-heuristics, Maximum-weight independent set}

\newcommand{\BibTeX}{\rm B\kern-.05em{\sc i\kern-.025em b}\kern-.08em\TeX}

\begin{document}

\maketitle
\begin{figure}[ht] 
	\centering
	\includegraphics[width=\linewidth]{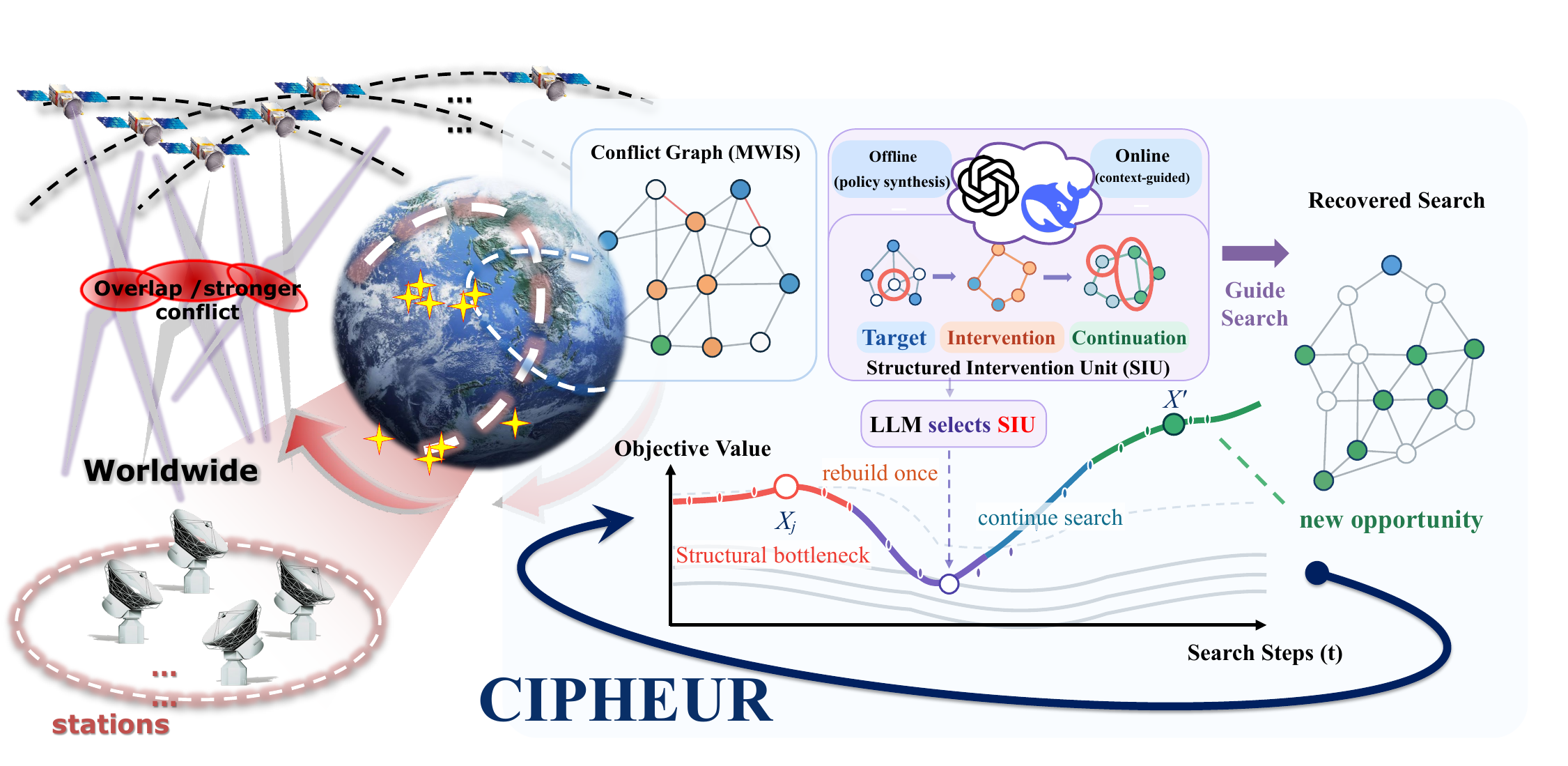}
	\caption{\textbf{CIPHEUR motivation.} Conflict evidence guides structural intervention and recovery in satellite--ground scheduling.}
	\Description{Satellite--ground contact conflicts form a weighted independent-set problem. Offline and online controllers select structural interventions that redirect an ongoing search, followed by recovery.}
	\label{fig:intro-motivation}
\end{figure}
\section{Introduction}
\label{sec:introduction}

Effective combinatorial optimization depends on matching search actions to the opportunities available in the current state. Learning-assisted optimization and selection hyper-heuristics have made such decisions an explicit target of algorithm design~\cite{bengio2021mlco,drake2020selection,gao2025neuralsolver}. Learning to schedule existing primal heuristics, for example, improves the allocation of search effort within a solver~\cite{chmiela2021scheduling}. In conflict-graph optimization, this allocation also has a structural dimension: the value of an operation depends on the activities it displaces and the combinations it makes accessible. Exploiting this relationship calls for control decisions that connect observed conflicts to the subsequent course of search.

We study maximum-weight independent set (MWIS) optimization on resource-labelled conflict graphs. Weighted vertices represent candidate activities, and edges encode pairwise incompatibility. In satellite--ground scheduling, resource and temporal labels describe the conflicts among communication opportunities. Two search trajectories can attain similar objective values while excluding different combinations of activities, making the choice of intervention target consequential. A reconstruction can also reduce a trajectory's immediate value while creating opportunities for later improvement or exchange. These dependencies motivate the joint choice of \emph{how to intervene, where to act, and how search should continue}. The resulting control problem links structural targeting with the delayed effects of intervention (Figure~\ref{fig:intro-motivation}).

Large language models (LLMs) offer complementary capabilities for addressing this problem. They can construct executable heuristics from problem descriptions and evaluative feedback~\cite{romera2024funsearch,liu2024eoh,ye2024reevo}, and make state-dependent heuristic choices during optimization~\cite{zhang2025laos,yang2025heuragenix}. Bringing these capabilities to conflict-graph search requires an actionable connection between contextual judgment and structural change. A useful decision must identify its target, organize the intervention and its continuation, and remain meaningful as the solver progresses. This motivates a common control abstraction through which both generated policies and runtime reasoning can direct a continuing search.

We propose \textbf{CIPHEUR}, an LLM-guided structural search-control framework for conflict-graph optimization. The framework translates conflict evidence and search experience into coordinated decisions about how to redirect an ongoing search. To express these decisions, we introduce the \textbf{Structured Intervention Unit (SIU)}, which jointly specifies a structural target, an intervention, and its continuation. Candidate SIUs are grounded in the conflicts affecting current search trajectories. The solution-quality changes observed across executed intervention--recovery periods inform subsequent choices, closing the loop between structural evidence and search behavior.

Two working modes realize control over this common SIU space. In the \emph{offline mode}, an LLM translates problem semantics and the control specification into customized executable policies. These policies select SIUs from evolving observations and may update their internal estimates from accumulated execution feedback. In the \emph{online mode}, an LLM makes the same type of SIU choice directly from current structural evidence and search history. Offline synthesis provides reusable control logic, while online reasoning retains contextual judgment during a solve. Both modes act through a shared execution interface that preserves the original objective and feasibility constraints; asynchronous online decisions allow native search to progress throughout model inference.

The contributions are threefold. \emph{First}, we propose CIPHEUR, an LLM-guided structural search-control framework for conflict-graph optimization that turns conflict evidence and search experience into adaptive control of ongoing search. \emph{Second}, we design SIUs that couple structural targets with intervention and recovery, using feedback from executed intervention--recovery periods to guide subsequent choices. \emph{Third}, we develop two policy realizations---offline controller synthesis and online contextual decision-making---over a shared executable interface, with asynchronous online execution sustaining native search and both modes preserving the original objective and feasibility constraints.

Experiments on eight constraint views of one satellite–ground scheduling source, with five runs per view and a 900-second evaluation cap, establish the competitiveness of CIPHEUR across 11 configurations. CIPHEUR-online increases mean scheduled contact duration by 14.84\% over the greedy baseline GWMIN. It also outperforms CHILS in all 40 paired runs. A separately frozen, one-shot LLM-generated controller improves the mean objective over CHILS without runtime model calls, while matched mechanism studies and cross-source evaluations further characterize the framework’s effectiveness.

\section{Related Work}
\label{sec:related-work}

\subsection{Independent-Set Search and Structural Guidance}
\label{sec:related-1}

MWIS heuristics exploit solution structure to combine intensification with exploration. METAMIS uses efficient local moves and path relinking within a GRASP framework~\cite{dong2022metamis}, while CHILS combines iterated local search with cooperation based on differences among solutions~\cite{grossmann2025chils}. Learning-based graph optimization~\cite{li2026lore,canturk2026transfergraph} offers another form of structural guidance: Li et al.~\cite{li2018gcntreesearch} use graph-convolutional predictions to guide tree search for independent sets. These approaches share the goal of using graph and solution information to expose promising combinations~\cite{alkhouri2025mis}.

CIPHEUR studies this relationship at the level of search interventions. Resource-specific blocking evidence connects candidate activities to the trajectories that exclude them, providing a basis for choosing where to redirect search. A coordinated continuation then explores the opportunities created by the intervention. The emphasis is on converting structural evidence into decisions about the course of a population search, complementing work on local moves, solution exchange, and vertex-level guidance.

\begin{figure*}[t]
	\centering
	\includegraphics[width=0.8\textwidth]{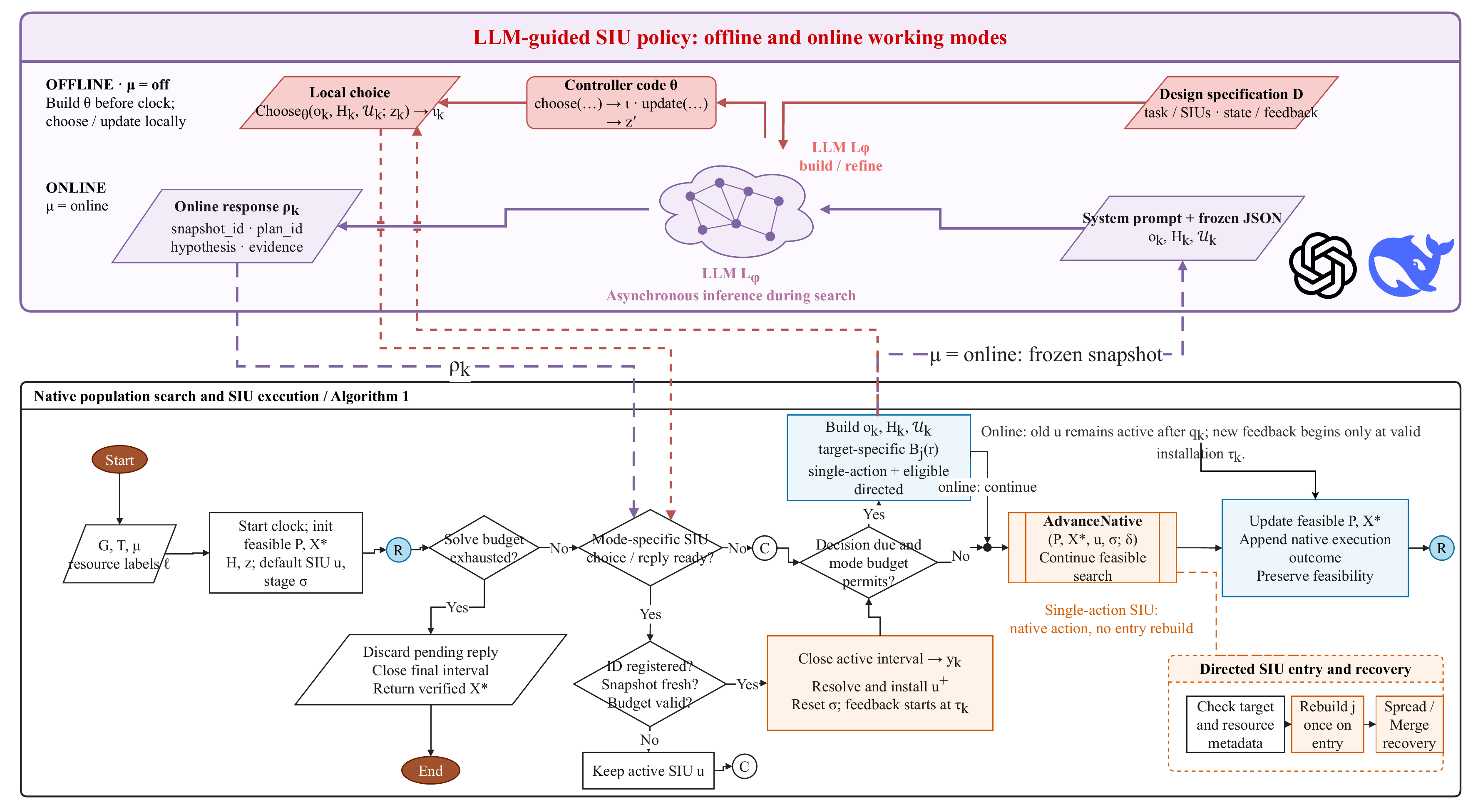}
	\caption{CIPHEUR overview. Both LLM modes select from one SIU registry; native search continues during online inference.}
	\Description{Four stages link structural observation, SIU construction, alternative policy modes and execution. A separate native-search lane continues during asynchronous online inference.}
	\label{fig:overview}
\end{figure*}

\subsection{Adaptive and Temporally Extended Search Control}
\label{sec:related-2}

Selection hyper-heuristics adapt the use of low-level heuristics to search state and history~\cite{drake2020selection}; dynamic algorithm configuration learns policies for changing algorithm settings during execution~\cite{adriaensen2022dac}. Learned heuristic scheduling similarly allocates effort among existing solver components~\cite{chmiela2021scheduling}. Adaptive large neighborhood search coordinates competing removal and insertion heuristics through performance feedback~\cite{ropke2006alns}, and learned neighborhood selection chooses variables for solver-based reoptimization~\cite{wu2021lns,ye2025llmlns,jaziri2026glns}. Together, these studies establish policy adaptation, effort allocation, and intervention targeting as complementary dimensions of search control~\cite{verdu2025onlineimprovement,ye2025anytime}.

Temporal abstraction provides a further dimension by treating an extended course of action as a decision unit, as formalized by the options framework~\cite{sutton1999options}. SIUs combine structural targeting with an explicit intervention--recovery sequence. A conflict anchor guides reconstruction priorities across the candidate activities, and the accompanying continuation specifies how the altered trajectory contributes to further search. Feedback summarizes quality changes over the executed sequence. This coupling makes the structural scope and observed effects of an intervention jointly available to the controller.

\subsection{LLM-Based Algorithm Design and Search Decisions}
\label{sec:related-3}

LLM-based algorithm design couples program generation with empirical evaluation~\cite{chen2026a2dept,vanstein2025llamea}. FunSearch evolves executable programs using systematic scoring~\cite{romera2024funsearch}; EoH jointly evolves heuristic ideas and code~\cite{liu2024eoh,yao2025meoh}; and ReEvo uses reflective feedback to guide this search~\cite{ye2024reevo,wu2026refineevo,malik2026pyvrpplus}. MCTS-AHD broadens exploration of heuristic programs through tree search~\cite{zheng2025mctsahd}, while EoH-S evolves complementary sets of heuristics~\cite{liu2026eohs}. These methods retain LLM design knowledge in reusable heuristic programs and portfolios~\cite{qiu2026e2oc,gungordu2026pathwise,kiet2026motif}.

A related line places model decisions within optimization~\cite{jiang2025llmco,chen2026dragon,jiang2026adcllm}. LAOS combines search-state information, operator credit, and replayed experience for adaptive operator selection~\cite{zhang2025laos}. LLMOA constructs sequences of low-level operators for continuous optimization~\cite{zhong2025llmoa}. HeurAgenix combines heuristic evolution with model-guided filtering and Monte Carlo continuation estimates for selection, and also trains a lightweight selector~\cite{yang2025heuragenix}.

CIPHEUR makes the joint choice of target, intervention, and recovery the common task for generated policies and online LLMs. The design focus is a control policy over structurally grounded interventions: directed SIUs specify which trajectory to redirect, which conflict relationships to emphasize, and how search should proceed afterward. Offline synthesis turns this decision task into a reusable controller, while online inference resolves it from the current search context. Both modes therefore operate over the same structural and temporal decision space.

Feedback relates policy decisions to changes in solution quality across their intervention--recovery periods. In online mode, asynchronous inference lets the optimizer continue searching while the model chooses its next intervention. The resulting design principle is to make structurally targeted search behavior the adaptable object, while retaining objective evaluation and feasibility enforcement in the native solver.

\section{CIPHEUR}
\label{sec:method}

CIPHEUR uses LLM capabilities to construct search controllers offline or choose interventions online. Both modes operate through a common loop: structural observation, SIU selection, native execution, and feedback (Figure~\ref{fig:overview}). The work mode is fixed before a solve.

\subsection{Problem and Control State}
\label{sec:control-problem}

Let $G=(V,E,w,\ell)$ be a resource-labelled conflict graph with positive weights $w$ and metadata $\ell$. Its feasible family and objective are
\begin{equation}
 \begin{aligned}
 \mathcal I(G)&=\{X\subseteq V:\ \{u,v\}\nsubseteq X\ \forall\{u,v\}\in E\},\\
 F(X)&=\sum_{v\in X}w_v .
 \end{aligned}
 \label{eq:mwis}
\end{equation}
Each scheduling vertex is a complete contact $v=(s_v,a_v,b_v,e_v)$ with integer duration $w_v=e_v-b_v$. For $u\prec v$ in start-time order, let $d_{uv}=b_v-e_u$. Contacts conflict when $(a_u=a_v\land d_{uv}<\delta^A_{a_u})$ or $(s_u=s_v\land d_{uv}<\delta^S_{s_u})$; equality with a threshold is feasible. Resource/time labels form $\ell$, and $G$ remains fixed throughout a solve.

The native state contains $m=4$ persistent trajectories $\mathcal P_t=(X_{t,j})_{j=0}^{m-1}$ and a historical incumbent $X_t^*$. The initial feasible set is replicated with distinct random states; each trajectory retains its solution and internal search state. We use CHILS local-search kernels~\cite{grossmann2025chils} within the CIPHEUR backend and report $F(X_T^*)$ at budget $T$.

At decision opportunity $k$, the common control task is
\begin{equation}
 u_k=\pi(o_k,\mathcal H_k,\mathcal U_k;z_k),\qquad u_k\in\mathcal U_k,
 \label{eq:control-policy}
\end{equation}
where $o_k$ is the current observation, $\mathcal H_k$ contains recent execution records, $\mathcal U_k$ is the SIU registry, and $z_k$ is optional policy memory. Observation time $q_k$ and installation time $\tau_k$ are distinct: online inference can overlap continued execution of the preceding SIU.

\begin{figure}[t]
	\centering
	\includegraphics[width=\columnwidth]{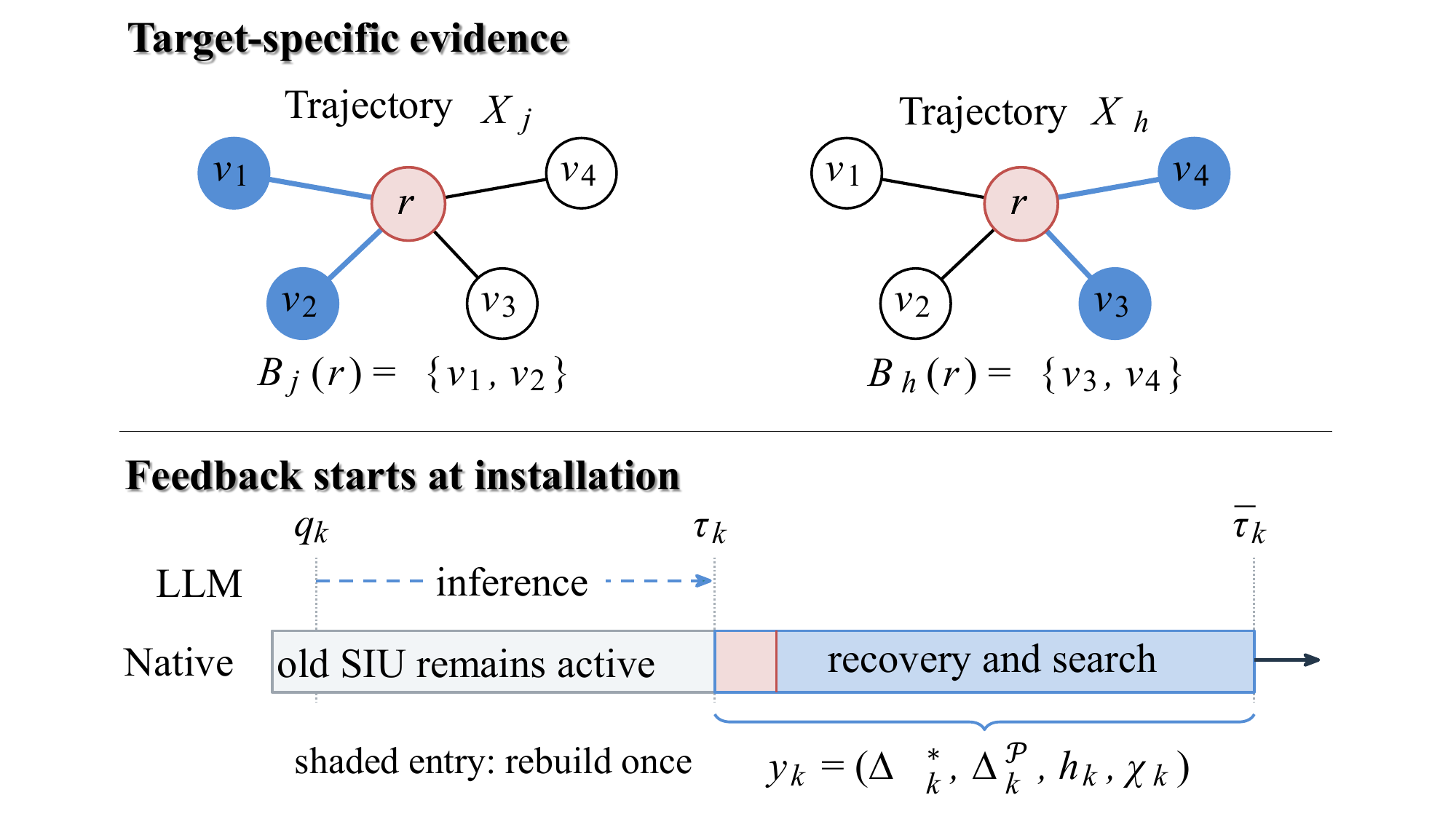}
	\caption{SIU grounding. (a) One anchor has different blockers across trajectories. (b) Online inference overlaps the old SIU; new feedback starts at installation $	au_k$.}
	\Description{The same unselected anchor has different selected blockers in two feasible trajectories. A timeline shows old-plan execution during inference and a new feedback interval beginning only at installation.}
	\label{fig:siu-boundaries}
\end{figure}

\subsection{From Conflict Evidence to SIUs}
\label{sec:siu}

\paragraph{Target-specific evidence.}
For anchor $r$ and trajectory $j$, selected blockers and their weight are
\begin{equation}
 B_j(r)=N_G(r)\cap X_{q_k,j},\qquad
 W_j(r)=\sum_{v\in B_j(r)}w_v .
 \label{eq:blockers}
\end{equation}
The observer records their counts, weights and resource relationships, together with the target's value gap, overlap with the population best, and whether $r$ is already selected. Figure~\ref{fig:siu-boundaries}(a) illustrates why evidence is recomputed for each target.

The observer considers the $L=128$ largest-weight vertices, retains those excluded and blocked by the population best, and ranks them by $w_r/\max\{1,W_{j_{\mathrm{best}}}(r)\}$. The first $R=6$ define the anchor set $\mathcal R_k$. Counts and weights use complete blocker sets; displayed blocker identities are capped at 16 and shared-blocker measures use these samples. Donor cross-conflicts sample up to 64 donor-exclusive vertices. Pairwise overlap is Jaccard similarity; population diversity is $|\bigcup_jX_j\setminus\bigcap_jX_j|/|\bigcup_jX_j|$. Progress, stagnation and resource use complete $o_k$. The anchor ratio ranks observations; executed outcomes evaluate plans.

\paragraph{Joint intervention representation.}
An SIU has the form
\begin{equation}
 u=(\iota,\omega,\mathcal S),\quad \omega=(j,r,c),\quad
 \mathcal S=((a_1,d_1),\ldots,(a_{n_u},d_{n_u})),
 \label{eq:siu}
\end{equation}
where $\iota$ is its identifier, $\omega$ binds the trajectory and anchor to resource-priority template $c$, and $\mathcal S$ specifies actions with nominal dwell times. Directed SIUs reconstruct $j$ once on entry, then continue with \textsc{Spread} or \textsc{Merge}: the former distributes search across trajectories, while the latter emphasizes exchange with a structurally different donor. Single-action SIUs have $\omega=\varnothing$.

Eligible targets $\mathcal J_k$ exclude slot zero and the population best. Given templates $\mathcal C$ and continuations $\mathcal A_{\mathrm{rec}}$, we construct
\begin{equation}
 \begin{aligned}
 \mathcal U_k=\mathcal U_{\mathrm{single}}\ \cup\ &\{u(j,r,c,a):j\in\mathcal J_k,\ r\in\mathcal R_k,\\
 &c\in\mathcal C,\ a\in\mathcal A_{\mathrm{rec}}\}_{\mathrm{eligible}} .
 \end{aligned}
 \label{eq:registry}
\end{equation}
Eligibility requires valid targets, target-specific evidence and the resource identities required by $c$. Descriptors are frozen with their observation for both policy modes. The nine single-action alternatives are \texttt{exploit}, \texttt{spread}, \texttt{merge}, \texttt{relay}, \texttt{antenna}, \texttt{satellite}, \texttt{kick}, \texttt{reseed}, and the native \texttt{adaptive} fallback. They cover intensification, diversification, exchange, resource-guided moves, entry perturbation and ordinary reseeding.

\paragraph{Reconstruction and continuation.}
Let $I_A(v,r)$ and $I_S(v,r)$ indicate shared resources and $t(v,r)=|b_v-b_r|$. For antenna, satellite and mixed templates, respectively, set
\begin{equation}
\begin{gathered}
 (M_A,g_A)=(1+2I_A,\delta^A_{a_v}),\; (M_S,g_S)=(1+2I_S,\delta^S_{s_v}),\\
 (M_M,g_M)=(1+I_A+I_S,\max\{\delta^A_{a_v},\delta^S_{s_v}\}),\\
 p_c(v;r)=\operatorname{sdiv}\!\left(w_v M_c,\operatorname{sdiv}(t(v,r),w_v+g_c)\right).
\end{gathered}
\label{eq:priority}
\end{equation}
Here $\operatorname{sdiv}(a,b)=\operatorname{trunc}_0[a/(1+|b|)]$; arithmetic saturates at magnitude $2^{61}-1$. All vertices are sorted by $p_c$, with seeded tie-breaking, then inserted greedily when compatible. Only a completed feasible construction replaces the target; interruption preserves the old set and ordering random state. Original weights determine $F$.

Directed SIUs use 8\,s for entry/search and 16\,s for recovery. The one entry reconstruction receives $\min(0.25,\delta/2,\text{remaining})$ seconds and at most $10^7$ work units within a native quantum $\delta$. Remaining time continues population search. The final action persists until replacement or termination. During donor exchange, independent candidates $A$ compete with $B=N(A)\cap X_j$ in a weighted bipartite subproblem, preserving $X_j\setminus B$. A completed flow solves this fixed exchange; bounded interruption retains feasibility. Accepted exchanges have nonnegative original-objective gain.

\subsection{Feedback over Executed Interventions}
\label{sec:feedback}

An interval starts at installation $\tau_k$ and closes at the next local decision checkpoint, accepted online replacement, fallback, or termination $\bar\tau_k$. With $Q_t=\sum_j F(X_{t,j})$, its outcome is
\begin{equation}
 \begin{gathered}
 y_k=(\Delta_k^*,\Delta_k^{\mathcal P},h_k,\chi_k),\; h_k=\bar\tau_k-\tau_k,\\
 \Delta_k^*=F(X_{\bar\tau_k}^*)-F(X_{\tau_k}^*),\;
 \Delta_k^{\mathcal P}=Q_{\bar\tau_k}-Q_{\tau_k}.
 \end{gathered}
 \label{eq:outcome}
\end{equation}
where $\chi_k$ records stages reached and execution status. Incumbent gain measures objective progress; population change exposes disruption and recovery. The record retains the initiating observation, while its gain and duration start at actual installation (Figure~\ref{fig:siu-boundaries}(b)).

Recent records and native-progress summaries inform subsequent decisions. Ongoing intervals carry an explicit unfinished status. Offline numerical controllers transform incumbent gain into a log-scaled improvement rate; online LLMs receive the underlying gains, durations and context. These are alternative uses of the same executed outcomes, with final evaluation governed by Eq.~\eqref{eq:mwis}.

\subsection{Two LLM Policy Realizations}
\label{sec:llm-modes}

\begin{figure}[t]
 \centering
 \includegraphics[width=\columnwidth]{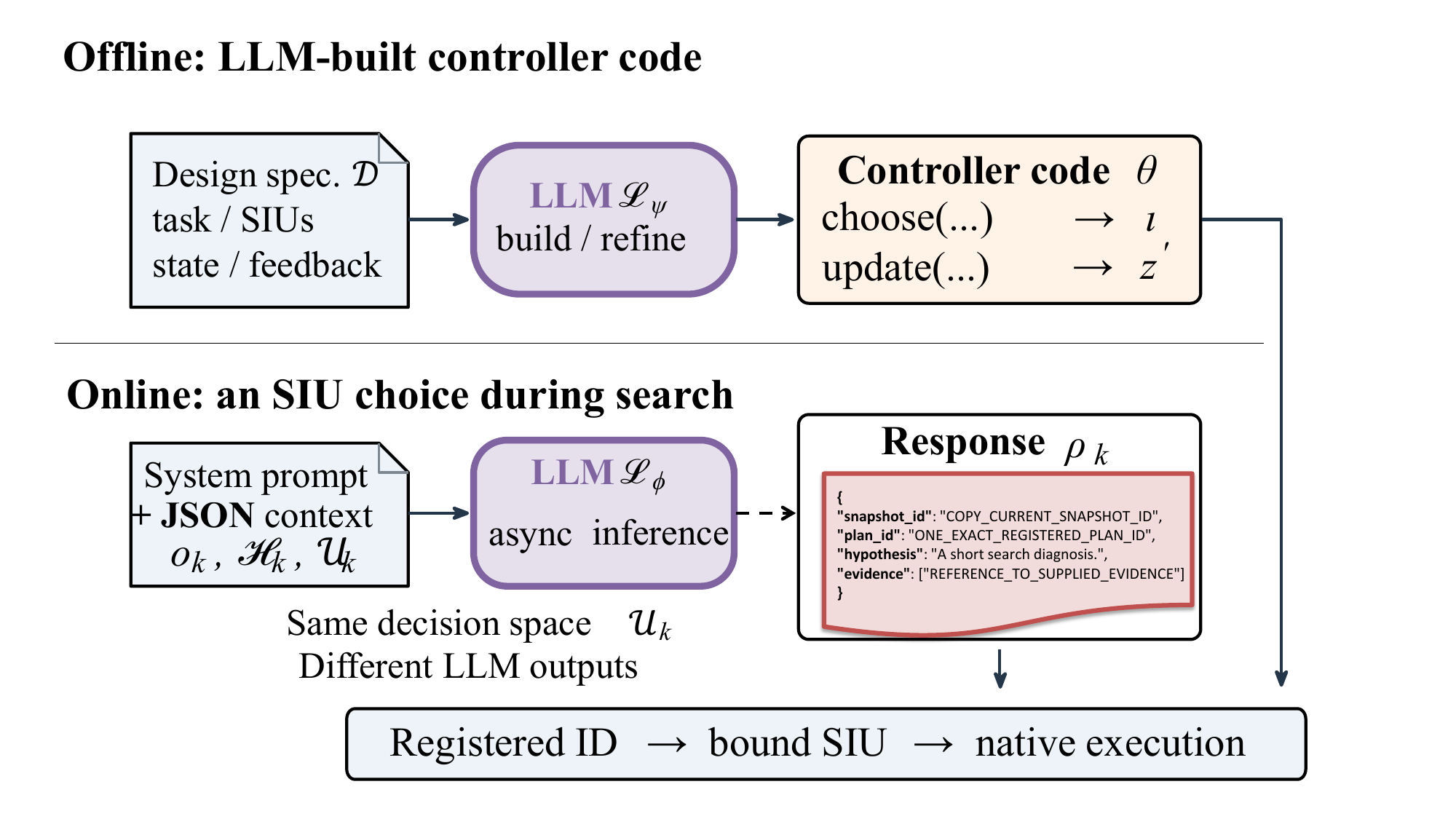}
 \caption{Offline controller construction and online SIU selection. Both modes share the native executor.}
 \Description{The offline LLM turns a design specification into choice and update code. The online LLM takes system instructions and current JSON context and returns snapshot ID, plan ID, hypothesis and evidence. Both outputs lead to shared native execution.}
 \label{fig:llm-modes}
\end{figure}

\paragraph{Offline: LLM construction of executable control.}
Before deployment, a design LLM $\mathcal L_\psi$ receives specification $\mathcal D$: the MWIS objective, observation semantics, SIU interface and feedback contract. Construction produces program $\theta$, with optional runtime memory $z_k$:
\begin{equation}
 \begin{aligned}
 \theta&\gets\textsc{ConstructWithLLM}(\mathcal L_\psi,\mathcal D),\\
 u_k&=\textsc{Choose}_{\theta}(o_k,\mathcal H_k,\mathcal U_k;z_k),\\
 z_{k+1}&=\textsc{Update}_{\theta}(z_k,o_k,u_k,y_k).
 \end{aligned}
 \label{eq:offline}
\end{equation}
Construction precedes the timed solve; choice and optional updates execute locally (Figure~\ref{fig:llm-modes}(a)), with memory reset between runs. The recorded one-shot realization is \texttt{off-GEN}. Archived \texttt{off-EXP} (KNN) and \texttt{off-VAL} (LinUCB) implement the same interface, separately from that generation record.

For these numerical realizations, let $f_k^{\rm obs}$ be the initiating incumbent and $x=\phi(o_k,u)\in\mathbb R^{59}$. Features comprise 10 global statistics, 15 action/template indicators, 10 target quantities and 24 interactions; identifiers are excluded. Feedback is
\begin{equation}
 r_k=\log\!\left(1+\frac{10^6\Delta_k^*}{\max(1,f_k^{\rm obs})\max(0.01,h_k)}\right).
 \label{eq:numerical-credit}
\end{equation}
EXP keeps the latest 160 samples, uses five nearest neighbors with $a_i=(0.2+\|x-x_i\|_2)^{-1}$, and scores $\sum_i a_ir_i/\sum_i a_i+0.25/\sqrt{1+\sum_i a_i}$. VAL starts from $A=I,b=0$, updates $A\gets A+xx^\top$, $b\gets b+r_kx$, and scores $b^\top A^{-1}x+0.25\sqrt{x^\top A^{-1}x}$. Both start with \textsc{Spread} and retain its preference $0.1/(1+n)$ after $n$ updates. Nonfinite features map to bounded neutral values.

\paragraph{Online: LLM reasoning over current SIUs.}
A fixed instruction $s_{\mathrm{sys}}$ and the current control context form the prompt; the model returns a data-only choice:
\begin{equation}
 \begin{aligned}
 p_k&=\textsc{Prompt}(s_{\mathrm{sys}};o_k,\mathcal H_k,\mathcal U_k),\\
 \rho_k&\gets\mathcal L_{\phi}(p_k),\qquad
 u_k=\textsc{Resolve}(\rho_k,\mathcal U_k).
 \end{aligned}
 \label{eq:online}
\end{equation}
The instruction asks the LLM to compare continued search with structural alternatives using progress, diversity, blocking evidence, recovery outcomes and remaining time. The JSON packet includes bounded summaries and previous choices; repeated SIU fields use shared tables. Figure~\ref{fig:llm-modes}(b) shows the response contract: the snapshot and plan IDs determine the executable SIU, while the diagnosis and evidence references are recorded as context. \textsc{Resolve} returns the registered descriptor for a correctly bound choice and $\bot$ otherwise. The full graph and selected sets remain native. Each fresh two-message request includes at most 12 action episodes, eight completed SIUs and eight previous choices, plus progress windows of 10, 40 and 120\,s. JSON factorization retains every registered choice. Explanatory fields are recorded as context; the selected registry identity controls execution.

\subsection{Asynchronous Execution and Complete Procedure}
\label{sec:execution}

Algorithm~\ref{alg:cipheur} realizes Eq.~\eqref{eq:control-policy} through the LLM-built program or online model; predeployment construction is outside the timed solve. Its local branch closes the preceding interval before updating and selecting; its online branch retains a frozen registry while native search continues. Returned choices undergo snapshot/descriptor checks, a 60\,s freshness limit and remaining-budget checks; request timeout is 45\,s. An idle inference lane and over 47\,s remaining are required to submit. The main thread alone installs plans and owns native state.

\begin{algorithm}[t]
\caption{CIPHEUR: LLM-guided SIU control}
\label{alg:cipheur}
\footnotesize
\begin{algorithmic}[1]
\Require $G,T,\mu\in\{\mathrm{off},\mathrm{online}\}$; quantum $\varepsilon$
\Statex \hspace{\algorithmicindent} Specification $\mathcal D$; instruction $s_{\mathrm{sys}}$; LLMs $\mathcal L_\psi,\mathcal L_\phi$
\Ensure Best verified independent set $X^*$
\If{$\mu=\mathrm{off}$}
    \State $\theta\gets\textsc{ConstructWithLLM}(\mathcal L_\psi,\mathcal D)$ \hfill\eqref{eq:offline}\label{line:llm-construct}
\EndIf
\State Start clock; initialize feasible $\mathcal P,X^*,\mathcal H,z$
\State Install default SIU $u$; reset stage $\sigma$ \hfill\eqref{eq:siu}
\While{elapsed time $<T$}
    \If{$\mu=\mathrm{online}$ and LLM reply $\rho_k$ is ready}
        \State $u^+\gets\textsc{Resolve}(\rho_k,\mathcal U_k)$ \hfill\eqref{eq:online}\label{line:resolve}
        \If{$u^+\neq\bot$ and freshness/budget checks pass}
            \State Close and record interval \hfill\eqref{eq:outcome}
            \State Install $u^+$; reset $\sigma$ to entry stage
        \EndIf
        \State Release the LLM inference lane
    \EndIf
    \If{a decision is due and mode-specific budgets allow}
        \If{$\mu=\mathrm{off}$}
            \State Close interval; update $z$ \hfill\eqref{eq:outcome},\eqref{eq:offline}\label{line:credit}
            \State Build $o_k,\mathcal H_k,\mathcal U_k$ \hfill\eqref{eq:blockers},\eqref{eq:registry}
            \State $u_k\gets\textsc{Choose}_{\theta}(o_k,\mathcal H_k,\mathcal U_k;z)$ \hfill\eqref{eq:offline}\label{line:local-choice}
            \State Validate/install $u_k$; reset $\sigma$; advance schedule
        \ElsIf{the LLM inference lane is idle}
            \State Build and freeze $o_k,\mathcal H_k,\mathcal U_k$ \hfill\eqref{eq:blockers},\eqref{eq:registry}
            \State $p_k\gets\textsc{Prompt}(s_{\mathrm{sys}};o_k,\mathcal H_k,\mathcal U_k)$ \hfill\eqref{eq:online}\label{line:prompt}
            \State Launch $\rho_k\gets\mathcal L_\phi(p_k)$ asynchronously\label{line:llm-call}
            \State Retain $\mathcal U_k$; advance request schedule
        \EndIf
    \EndIf
    \If{the solve budget is exhausted} \State \textbf{break} \EndIf
    \State Advance due stages in $\sigma$; retain the final action
    \State $\delta\gets\min\{\varepsilon,\text{stage remainder},\text{solve remainder}\}$
    \State $\textsc{AdvanceNative}(\mathcal P,X^*,u,\sigma;\delta)$\label{line:native}
    \State Append the native execution outcome
\EndWhile
\State Discard pending replies; close the final interval \hfill\eqref{eq:outcome}
\State \Return $X^*$ after original-graph and objective checks \hfill\eqref{eq:mwis}
\end{algorithmic}
\end{algorithm}

The LLM steps are explicit: line~\ref{line:llm-construct} produces controller code; lines~\ref{line:prompt}--\ref{line:llm-call} form and submit an online decision; line~\ref{line:resolve} maps the reply to a frozen SIU. Native advancement at line~\ref{line:native} proceeds while inference is pending. Before a targeted entry, the executor rechecks slot protection and metadata. Rejection advances to the registered continuation, or installs the default plan when none remains. It retains the requested target's identity throughout. Interval closure uses Eq.~\eqref{eq:outcome}; policy updates skip the initial fallback interval and occur only when feedback is available.

\paragraph{Feasibility and cost.}
Starting from feasible states, native moves preserve independence and reconstruction commits only completed compatible sets. The incumbent accepts only a better feasible set. Induction over completed operations therefore preserves population feasibility and nondecreasing incumbent value, while allowing individual trajectories to lose value during diversification. With $p$ eligible targets, $R$ anchors, $C$ templates, $A$ continuations and $S$ single-action plans, registry size is at most $S+pRCA$ (117 in this realization). Full reconstruction costs $O(|V|\log|V|+|E|)$ before work/time limits. Observation and local execution count toward the solve; online inference overlaps native search. Checks between bounded operations yield a soft wall-clock budget. All decisive observation, priority, feedback and execution choices are specified above; full helper code accompanies the submission.

\section{Experiments}
\label{sec:experiments}
We examine terminal quality, the development of online gains, and the contributions of structural control. The historical comparison, fresh matched screen, completed control extensions, and new-source tests remain separate evaluation frames.

\subsection{Instances, Comparators and Protocol}
\label{sec:exp-protocol}
CP-SCALE-AU-L002 supplies eight views of the same contacts and weights. Homogeneous views $H_a$ use antenna gaps of 340, 680, 1200 and 1800\,s; mixed views $M_{a/b}$ use western/eastern gaps $(340,1200)$, $(680,1200)$, $(1200,340)$ and $(1200,680)$\,s. Each view uses seeds 67, 71, 73, 79 and 83 under a 900\,s outer wall cap. The registered CIPHEUR/CHILS campaign matches four trajectories and one native thread per solve. Online requests begin at 20\,s, recur at 40\,s opportunities and have a cap of 21; the archived model identifier is \texttt{gpt-6-luna}. Model inference uses additional compute. Observation and local selection consume solve time; inference overlaps native search.

Table~\ref{tab:main} retains all 11 completed configurations. External methods include CHILS-p4-custom~\cite{grossmann2025chils}, Feasibility Jump~\cite{luteberget2023fj}, FastWVC~\cite{cai2019fastwvc}, GRASP~\cite{feo1995grasp}, StableSolver LS and LNS~\cite{stablesolver}, and GWMIN~\cite{sakai2003gwmin}. Four CIPHEUR realizations complete the comparison. LS and LNS ignore seed arguments and therefore use five fresh process repeats. LS has a 9\,GiB limit, requests 850\,s native time and records 856.8--874.4\,s wall time; LNS requests 890\,s and records 891.1--892.5\,s. GWMIN stops before the cap. All accepted results passed recorded graph-feasibility and integer-objective audits. External-table comparisons are descriptive; pairwise wins below use matched frames.

EXP and VAL are the archived numerical controls defined in Section~\ref{sec:llm-modes}; \emph{off} denotes zero runtime model calls. GEN follows a separate frozen generation protocol. The released EXP/VAL features clip elapsed and remaining times after division by 360\,s; that convention is retained in this 900\,s comparison. Means give equal weight to views after averaging their runs. For terminal comparison, $E_v=\max_{b\in\mathcal B}\bar F_{b,v}$ is the best mean among all seven external rows. Table~\ref{tab:main} shows $100(\bar F_{m,v}/E_v-1)$, absolute overall means and mean within-view ranks.

\begin{table*}[t]
\centering
\caption{L002 terminal quality (five runs/view). View entries are gains over $E_v$ (\%); $E_v$ and Mean use $10^3$ contact-s. Rank averages within-view ranks; bold marks bests. $^{\dagger}$GEN was separately frozen; LS/LNS use process repeats.}
\label{tab:main}
\small
\setlength{\tabcolsep}{2.5pt}
\begin{tabular*}{0.8\textwidth}{@{\extracolsep{\fill}}lrrrrrrrrrr@{}}
\toprule
Method & Mean$\uparrow$ & $H_{340}$ & $H_{680}$ & $H_{1200}$ & $H_{1800}$ & $M_{340/1200}$ & $M_{680/1200}$ & $M_{1200/340}$ & $M_{1200/680}$ & Rank$\downarrow$\\

$E_v$ ($10^3$ s) & --- & 1624.295 & 1246.823 & 884.035 & 684.860 & 1259.316 & 1064.712 & 1261.095 & 1062.766 & ---\\
\midrule
\multicolumn{11}{@{}l}{\textit{External baselines}}\\
CHILS & 1,134.943 & 0.00 & 0.00 & -0.63 & -0.41 & 0.00 & 0.00 & 0.00 & 0.00 & 5.38 \\
\addlinespace[1pt]
Feasibility Jump~\cite{luteberget2023fj} & 1,131.650 & -0.06 & -0.25 & -1.40 & \textbf{0.00} & -0.31 & -0.59 & -0.27 & -0.43 & 6.12 \\
\addlinespace[1pt]
StableSolver LS~\cite{stablesolver} & 1,131.306 & -0.80 & -0.62 & 0.00 & -0.61 & -0.28 & -0.26 & -0.38 & -0.14 & 6.38 \\
\addlinespace[1pt]
StableSolver LNS~\cite{stablesolver} & 1,125.482 & -1.08 & -1.05 & -0.81 & -0.63 & -0.85 & -0.98 & -1.01 & -0.75 & 8.38 \\
\addlinespace[1pt]
FastWVC~\cite{cai2019fastwvc} & 1,124.488 & -1.41 & -1.59 & -0.20 & -0.02 & -1.14 & -0.94 & -1.19 & -0.75 & 7.38 \\
\addlinespace[1pt]
GRASP~\cite{feo1995grasp} & 1,057.130 & -6.21 & -8.68 & -4.77 & -10.95 & -6.00 & -7.19 & -6.21 & -7.00 & 10.00 \\
\addlinespace[1pt]
GWMIN~\cite{sakai2003gwmin} & 991.619 & -12.08 & -14.54 & -8.46 & -28.54 & -10.24 & -11.62 & -10.74 & -11.20 & 11.00 \\
\addlinespace[1pt]
\midrule
\multicolumn{11}{@{}l}{\textit{CIPHEUR family}}\\
CIPHEUR-off-EXP & 1,138.452 & \textcolor{CIPblue}{+0.13} & \textcolor{CIPblue}{+0.23} & \textbf{+0.24} & -0.19 & \textcolor{CIPblue}{+0.25} & \textcolor{CIPblue}{+0.36} & \textcolor{CIPblue}{+0.13} & \textbf{+0.49} & 2.50 \\
\addlinespace[1pt]
CIPHEUR-off-VAL & 1,138.418 & \textcolor{CIPblue}{+0.13} & \textcolor{CIPblue}{+0.23} & \textcolor{CIPblue}{+0.09} & -0.18 & \textcolor{CIPblue}{+0.27} & \textbf{+0.38} & \textcolor{CIPblue}{+0.20} & \textcolor{CIPblue}{+0.47} & 2.50 \\
\addlinespace[1pt]
CIPHEUR-off-GEN$^{\dagger}$ & 1,136.347 & \textcolor{CIPblue}{+0.03} & \textcolor{CIPblue}{+0.14} & -0.21 & -0.31 & \textcolor{CIPblue}{+0.05} & \textcolor{CIPblue}{+0.11} & -0.04 & \textcolor{CIPblue}{+0.31} & 4.62 \\
\addlinespace[1pt]
\rowcolor{CIPblue!8}
\textbf{CIPHEUR-online} & \textbf{1,138.760} & \textbf{+0.17} & \textbf{+0.24} & \textcolor{CIPblue}{+0.22} & -0.15 & \textbf{+0.32} & \textcolor{CIPblue}{+0.36} & \textbf{+0.21} & \textcolor{CIPblue}{+0.48} & \textbf{1.75} \\
\addlinespace[1pt]
\bottomrule
\end{tabular*}
\end{table*}

\subsection{Final Quality and Late-Search Improvement}
\label{sec:exp-quality}
\paragraph{Competitiveness across constraint views.}
CIPHEUR-online attains the highest mean, 1,138,760.369 contact-seconds, and best average rank, 1.75. Online, EXP and VAL each exceed the strongest external mean in seven views. CHILS defines that reference in six views, LS in $H_{1200}$ and Feasibility Jump in $H_{1800}$; online trails the latter by 0.148\%. Against CHILS alone, online gains 3,817.583 contact-seconds (0.336\%) and wins all 40 graph--seed pairs. Its additional mean gains over EXP and VAL are 308.457 and 342.692 seconds, with 24/40 wins against each: a smaller, variable improvement within the common framework.

\begin{figure}[t]
 \centering
 \includegraphics[width=\columnwidth]{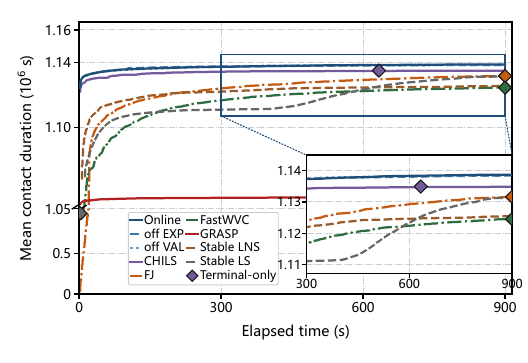}
 \caption{L002 progress (40 runs). The main axis uses 20:1 compression below $1.05\times10^6$ contact-s; inset: 300--900\,s. CHILS, FJ and FastWVC are separate reruns.}
 \Description{A full-range plot contains an internal 300--900 second enlargement. Solid method-colored diamonds mark terminal-only comparison means; lines show recorded incumbent trajectories.}
 \label{fig:search-progress-zoom}
\end{figure}

Figure~\ref{fig:search-progress-zoom} shows the recorded method trajectories and a late-search enlargement. The horizontal coordinate follows each method's recorded solver or outer clock. Filled diamonds identify frozen terminal-only means; their positions mark the recorded end or budget cap, not first attainment.

\paragraph{Same-run improvement and attainment.}
For trajectories, $R_v$ is the best audited individual terminal value among the original online/EXP/VAL/CHILS runs in view $v$. Define the empirical-reference shortfall
\begin{equation}
 g_{m,v,s}(t)=100\frac{R_v-F_{m,v,s}(t)}{R_v},\qquad
 \bar g_m(t)=\frac{1}{40}\sum_{v,s}g_{m,v,s}(t).
 \label{eq:reference-gap}
\end{equation}
$R_v$ is a feasible reference, distinct from $E_v$ and from a certified optimum. Figure~\ref{fig:anytime-quality} uses saved epoch-end incumbents from the same 900\,s runs, displaying all 12 released checkpoints. CHILS supplies terminal values only.

At 360\,s, online has larger shortfall than EXP and VAL; by 900\,s this ordering reverses (Table~\ref{tab:late-phase}). Its $\bar g(360)-\bar g(900)$ improvement is \textbf{1.50 times EXP's and 1.86 times VAL's}, a ratio of quality gains rather than runtime speedups. Online attains the strictest positive target, 0.05\%, in 10/40 runs versus 6/40 for each comparator. The complete nine-threshold grid also preserves the exception at 0.4\% (online 38, EXP 39) and equal attainment at 0.6\% and 1\%. The recorded time-normalized 20--900\,s shortfall areas (AOC) are 0.24065\%, 0.24177\% and 0.25226\%: online's terminal advantage coexists with near-equal whole-run area against EXP.

\begin{table}[t]\centering\footnotesize
\caption{Late progress, attainment, and matched control effects. Control $\Delta$ is online minus control.}
\label{tab:late-phase}\label{tab:heterogeneity}
\setlength{\tabcolsep}{2.4pt}
\begin{tabular*}{\columnwidth}{@{\extracolsep{\fill}}lrrrr@{}}\toprule
\multicolumn{5}{@{}l}{\textit{Late shortfall/AOC (\%) and improvement (pp)}}\\
Mode & $\bar g_{360}\downarrow$ & $\bar g_{900}\downarrow$ & $\Delta_{360:900}\uparrow$ & AOC$\downarrow$\\\midrule
\rowcolor{CIPblue!8}
\textbf{Online} & 0.23746 & \textbf{0.12698} & \textbf{0.11048} & \textbf{0.24065}\\
Off-EXP & 0.22471 & 0.15113 & 0.07359 & 0.24177\\
Off-VAL & \textbf{0.21849} & 0.15914 & 0.05936 & 0.25226\\
\bottomrule\end{tabular*}\par\vspace{2pt}
{\scriptsize\setlength{\tabcolsep}{1.2pt}
\begin{tabular*}{\columnwidth}{@{\extracolsep{\fill}}l*{9}{r}@{}}\toprule
\multicolumn{10}{@{}l}{\textit{Terminal attainment (counts out of 40)}}\\
$\epsilon$ (\%) & 0 & 0.05 & 0.10 & 0.15 & 0.20 & 0.30 & 0.40 & 0.60 & 1.00\\\midrule
Online & \textbf{3} & \textbf{10} & \textbf{17} & \textbf{27} & \textbf{33} & \textbf{37} & 38 & \textbf{40} & \textbf{40}\\
Off-EXP & \textbf{3} & 6 & 12 & 23 & 31 & 36 & \textbf{39} & \textbf{40} & \textbf{40}\\
Off-VAL & 2 & 6 & 16 & 22 & 31 & 35 & 36 & \textbf{40} & \textbf{40}\\
\bottomrule\end{tabular*}\par}
\vspace{2pt}
\begin{tabular*}{\columnwidth}{@{\extracolsep{\fill}}lrrrr@{}}\toprule
\multicolumn{5}{@{}l}{\textit{Matched control split (contact-s; wins/6)}}\\
 & \multicolumn{2}{c}{Homogeneous} & \multicolumn{2}{c}{Mixed gaps}\\
Control & Mean $\Delta$ & Wins & Mean $\Delta$ & Wins\\\midrule
SIU-Off & +3,416.7 & 6/6 & +5,065.7 & 6/6\\
Fixed-Spread & +576.9 & 3/6 & +2,210.4 & 6/6\\
Random-SIU & -338.3 & 1/6 & +1,415.1 & 5/6\\
No-Struct & -797.0 & 2/6 & +1,060.3 & 6/6\\
Single-Op & -836.1 & 2/6 & +1,023.4 & 6/6\\
\bottomrule\end{tabular*}\end{table}

\subsection{Control Contributions and Observed Recovery}
\label{sec:exp-mechanism}
The fresh mechanism screen matches graph, seed and host partition on $H_{340}$, $H_{1800}$, $M_{340/1200}$ and $M_{1200/340}$, with three seeds each. SIU-Off retains the initial diversity-adaptive native plan throughout; Fixed-Spread keeps decision checkpoints but always selects \textsc{Spread}. Random-SIU samples legal units, Single-Op restricts the registry to single actions, and No-Struct masks resource witnesses and plan context while preserving legal IDs and bindings. Thus No-Struct retains cues in action identifiers. All use 900\,s and the same objective checks.

\begin{figure}[t]
 \centering
 \includegraphics[width=\columnwidth]{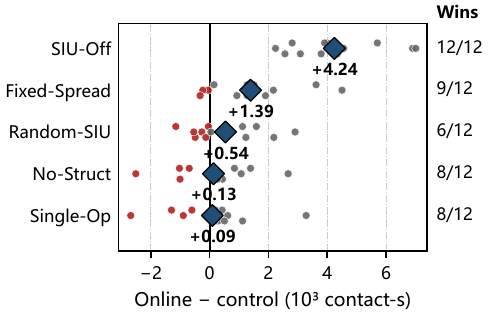}
 \caption{Matched mechanism screen. Each dot is an online-minus-control difference (12 pairs per control); diamonds mark means and the right column lists positive pairs.}
 \Description{All twelve paired differences favor online against SIU-Off. Nine favor online against Fixed-Spread, while the three finer comparisons include both positive and negative differences.}
 \label{fig:ablation}
\end{figure}

Figure~\ref{fig:ablation} retains every paired gain and loss. Full online control gains \textbf{4,241.212 contact-seconds over SIU-Off} (12/12 wins) and 1,393.654 over Fixed-Spread (9/12). Fixed-Spread changes both adaptivity and the action mix, so the result supports the complete control configuration. Finer differences are +538.377 for Random-SIU, +131.678 for No-Struct and +93.655 for Single-Op, with 6/12, 8/12 and 8/12 wins. These smaller, seed-dependent contrasts provide limited component-specific evidence. Separately completed 40-position controls average 1,134,403.037 (SIU-Off) and 1,137,567.142 (Fixed-Spread) seconds. Their matched online extension was incomplete; historical online runs are not substituted as partners.

\begin{figure*}[t]
	\centering
	\begin{minipage}[t]{.49\textwidth}\centering
		\includegraphics[width=0.8\linewidth]{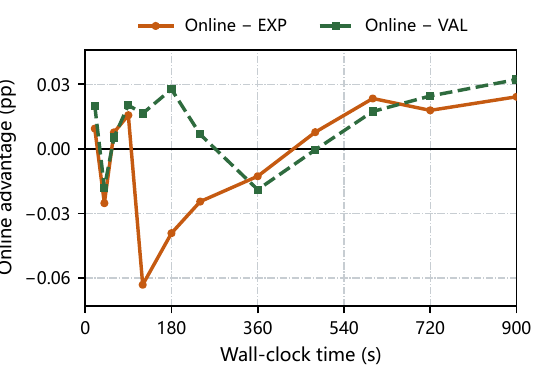}
	\end{minipage}\hfill
	\begin{minipage}[t]{.49\textwidth}\centering
		\includegraphics[width=0.8\linewidth]{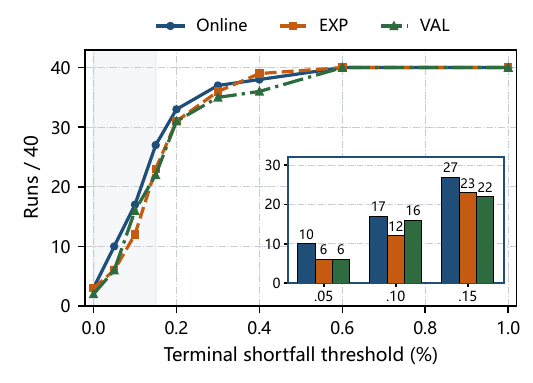}
	\end{minipage}
	\caption{Search progress and attainment. (a) Mean online-minus-programmatic quality at 12 saved checkpoints. (b) Counts across nine terminal thresholds (40 runs); inset: three strictest positive thresholds.}
	\Description{Online is slightly behind at the start of the late-search window and ahead at the final checkpoints. At 900 seconds it attains the strictest positive reference-gap target in ten runs, against six for each programmatic comparator.}
	\label{fig:anytime-quality}
	
	\centering
	\begin{minipage}[t]{.325\textwidth}\centering
		\includegraphics[width=\linewidth]{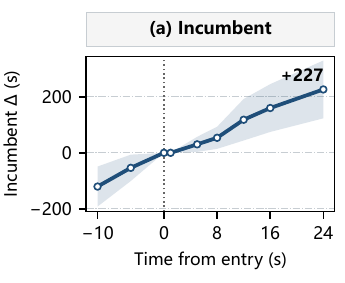}
	\end{minipage}\hfill
	\begin{minipage}[t]{.325\textwidth}\centering
		\includegraphics[width=\linewidth]{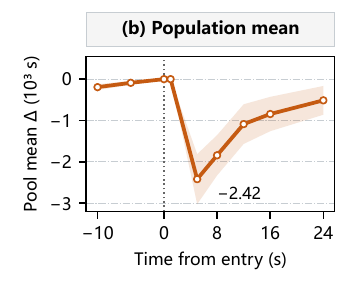}
	\end{minipage}\hfill
	\begin{minipage}[t]{.325\textwidth}\centering
		\includegraphics[width=\linewidth]{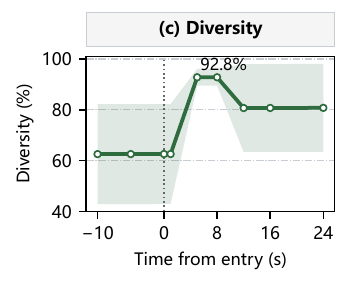}
	\end{minipage}
	\caption{Observed recovery after 401 directed reseeds. Curves average 40 run means; bands show descriptive $\pm1$ sample SD. Panel (b) is the population mean.}
	\Description{The historical incumbent increases, mean population value initially decreases and then recovers, and population diversity rises after reseeding.}
	\label{fig:recovery}
\end{figure*}

The bottom panel of Table~\ref{tab:late-phase} localizes the fine-control differences. On mixed-gap views, full control exceeds No-Struct by 1,060.3 seconds and Single-Op by 1,023.4 seconds, with 6/6 positive pairs for each. Both comparisons reverse in the homogeneous-group mean (2/6 positive). Structure evidence and coordinated recovery therefore have their clearest observed benefit in the mixed-constraint subset; the two-view groups support a descriptive explanation, not a general interaction claim.

\paragraph{A measured intervention--recovery pattern.}
Of 406 recorded directed intervals, 401 install a native reseed. Figure~\ref{fig:recovery} averages events within each run and then weights the 40 runs equally. At +5\,s, mean population value is 2,417.631 seconds below entry and diversity rises from 62.6\% to 92.8\%. At +24\,s, the deficit is 512.513 seconds: \textbf{78.8\% has recovered}, while the historical incumbent improves by 226.823 seconds. All 401 events remain installed through +24\,s; 400 remain at +30\,s. The population curve describes the pool mean, not only the targeted slot. These observational changes explain why immediate trajectory loss can coexist with later progress; they do not isolate the causal effect of reseeding alone.

\paragraph{Native work continues during inference.}
Among 840 online calls, 838 install a valid plan. Request-to-receipt latency has median 11.06\,s and 90th percentile 13.87\,s. Pending intervals total 9,845.261\,s and overlap 9,837.360\,s of native work: \textbf{99.92\% overlap}, or 245.9\,s per solve, with 2,720,049 native cycles. HTTP-only overlap is 99.99\%. These intervals lie inside the solve budget; they establish sustained execution during model latency, not an additive time saving or measured synchronous-control speedup.

\subsection{Offline Synthesis and Cross-Source Use}
\label{sec:exp-offline}
The prospective GEN test issued one recorded \texttt{gpt-6-luna} request at low reasoning effort for a deterministic \texttt{select\_plan(packet, plans)} function. Only the supplied state and legal descriptors were accessible. The first candidate was frozen without objective-guided revision. All 84 saved interface packets yielded legal IDs, deterministic repeats and unchanged inputs. These packets came from four L002 views at seed 83, so those four graph--seed positions overlap the subsequent 40-position evaluation; validation tested the interface rather than tuning objective performance. GEN completes all 40 runs with zero deployment model calls and a mean of 1,136,347.357 seconds: \textbf{+1,404.572 over CHILS (0.124\%)}. Its lower mean than EXP/VAL shows that a usable one-shot policy need not dominate established control realizations.

\begin{table}[t]\centering\footnotesize
\caption{L003 cross-source means ($10^3$ contact-s; three runs per view). Gains and matched wins are relative to CHILS.}
\label{tab:l003}
\setlength{\tabcolsep}{2.4pt}
\begin{tabular*}{\columnwidth}{@{\extracolsep{\fill}}lrrr@{}}\toprule
View & CHILS & Off-EXP & Off-VAL\\\midrule
AU-H & 1,623.290 & \textbf{1,627.711} & 1,627.468\\
AU-M & 1,259.639 & \textbf{1,264.050} & 1,263.209\\
AP-H & 1,690.812 & 1,692.549 & \textbf{1,692.806}\\
AP-M & 1,283.948 & \textbf{1,284.905} & 1,284.875\\\midrule
Mean$\uparrow$ & 1,464.422 & \textbf{1,467.304} & 1,467.090\\
Gain (\%) & --- & +0.197 & +0.182\\
Wins & --- & 11/12 & 12/12\\
\bottomrule\end{tabular*}\end{table}

L003 adds AU and AP regional contact sets, each under homogeneous and mixed constraints with three seeds.

The authors attest to excluding these sources from prompt design, tuning and controller selection. Frozen EXP/VAL programs use fresh per-solve memory. Table~\ref{tab:l003} brings per-view means into the main evaluation; the sample SDs remain in the supplement. EXP and VAL improve the overall CHILS mean by 0.197\% and 0.182\%, lead all four view means, and win 11/12 and 12/12 pairs. Only the three completed arms are included; online and GEN cross-source runs remain incomplete.

\paragraph{Scope.}
The evidence covers a few contact sources and a 12-position fine-control screen. Time-feature scaling remains a configuration constraint, L003 non-use is author-attested, and synchronous-control/online cross-source tests are unfinished. Missing token receipts prevent complete inference-cost accounting.

\section{Conclusion}
CIPHEUR treats conflict-graph search control as a joint decision over structural targets, interventions and their continuation. SIUs give generated programs and online LLMs a common executable task, while native search preserves the original optimization semantics. The completed experiments show broad terminal competitiveness, stronger late-search progress for the online configuration, and a reconstruction--recovery pattern that motivates interval-level feedback. A frozen one-shot controller further demonstrates deployment without runtime model calls. Together, these findings identify structural search control as a practical way to use LLM design knowledge and contextual reasoning in combinatorial optimization.

\bibliographystyle{ACM-Reference-Format}
\bibliography{CIPHEUR_references}

\end{document}